\documentclass{article}

\usepackage{arxiv}          % arXiv constructs
\usepackage[utf8]{inputenc} % allow utf-8 input
\usepackage[T1]{fontenc}    % use 8-bit T1 fonts
\usepackage{hyperref}       % hyperlinks
\usepackage{url}            % simple URL typesetting
\usepackage{booktabs}       % professional-quality tables
\usepackage{amsfonts}       % blackboard math symbols
\usepackage{nicefrac}       % compact symbols for 1/2, etc.
\usepackage{microtype}      % micro-typography
\usepackage{graphicx}	      % image figures
\usepackage{enumitem}	      % ordered lists
\usepackage{tabularx}       % auto-wide table columns
\usepackage{multirow}       % multirow table cells
\usepackage{caption}        % enhanced captioning
\usepackage{siunitx}        % scientific notation and SI unit
\usepackage[lofdepth,lotdepth]{subfig}
\usepackage[style=numeric,sorting=none]{biblatex}

\date{} % Removes the date that is present by default
\renewcommand{\undertitle}{} % Removes the tiny "A Preprint" subtitle
\renewcommand{\headeright}{} % Removes "A Preprint" from subsequent page headers

\newcommand{\rightparenthesis}{)} % chktex 9 chktex 10

\newenvironment{figure_in_table}{\captionsetup{type=figure}}{}

\hypersetup{
    colorlinks=true,
    urlcolor=blue,
    pdftitle={No Routing Needed Between Capsules},
    pdfpagemode=FullScreen,
}

\title{No Routing Needed Between Capsules}

\author{
  Adam~Byerly\\
  Department of Electronic and Electrical Engineering\\
  Brunel University London\\
  Uxbridge, UB8 3PH UK \\
  Department of Computer Science and Information Systems\\
  Bradley University\\
  Peoria, Il, 61615 USA\\
  \texttt{abyerly@fsmail.bradley.edu} \\
  \And{}
  Tatiana~Kalganova \\
  Department of Electronic and Electrical Engineering\\
  Brunel University London\\
  Uxbridge, UB8 3PH UK \\
  \texttt{tatiana.kalganova@brunel.ac.uk} \\
  \And{}
  Ian~Dear \\
  Department of Electronic and Electrical Engineering\\
  Brunel University London\\
  Uxbridge, UB8 3PH UK \\
  \texttt{ian.dear@brunel.ac.uk}
}

\begin{document}

\maketitle

\begin{abstract}
  Most capsule network designs rely on traditional matrix multiplication between capsule layers and computationally expensive routing mechanisms to deal with the capsule dimensional entanglement that the matrix multiplication introduces.  By using Homogeneous Vector Capsules (HVCs), which use element-wise multiplication rather than matrix multiplication, the dimensions of the capsules remain unentangled.  In this work, we study HVCs as applied to the highly structured MNIST dataset in order to produce a direct comparison to the capsule research direction of Geoffrey Hinton, et al.  In our study, we show that a simple convolutional neural network using HVCs performs as well as the prior best performing capsule network on MNIST using 5.5\(\times{}\) fewer parameters, 4\(\times{}\) fewer training epochs, no reconstruction sub-network, and requiring no routing mechanism.  The addition of multiple classification branches to the network establishes a new state of the art for the MNIST dataset with an accuracy of \textbf{99.87\%} for an ensemble of these models, as well as establishing a new state of the art for a single model (\textbf{99.83\%} accurate).
\end{abstract}


\section{Introduction and Related Work}\label{sec:introduction}

Capsules (vector-valued neurons) have become a more active area of research since~\cite{Sabour2017}, which demonstrated near state of the art performance on MNIST~\cite{Lecun2010} classification (at 99.75\%) by using capsules and a routing algorithm to determine which capsules in a previous layer feed capsules in the subsequent layer.  MNIST is a classic image classification dataset of hand-written digits consisting of 60,000 training images and 10,000 validation images.  Studying MNIST, due to the more highly structured content as compared to many other image datasets, allows for the use of more informed data augmentation techniques and, when using capsules, the ability to investigate the capsules' interpretability.  In~\cite{Hinton2018}, the authors extended their work by conducting experiments with an alternate routing algorithm.  Research in capsules has since focused mostly on various computationally expensive routing algorithms (\cite{Venkataraman2020}\cite{Amer2020}).  In~\cite{Byerly2019}, we proposed a capsule design that used element-wise multiplication between capsules in subsequent layers and relied on backpropagation to do the work that prior capsule designs were relying on routing mechanisms for.  We referred to this capsule design as homogeneous vector capsules (HVCs).  In this work, we directly extend the work of~\cite{Hinton2011} and~\cite{Sabour2017} on capsules applied to MNIST by applying HVCs to MNIST.\@ By using this capsule design, we avoid the the computationally expensive routing mechanisms of prior capsule work and we surpass the performance of~\cite{Sabour2017} on MNIST, all while requiring 5.5\(\times{}\) fewer parameters, 4\(\times{}\) fewer epochs of training, and using no reconstruction sub-network.

Many of the best performing convolutional neural networks (CNNs) of the past several years have explored multiple paths from input to classification~\cite{Szegedy2015a}\cite{Szegedy2015b}\cite{He2015}\cite{Zhou2020}\cite{Wang2020}\cite{Ciresan2012}.  The idea behind multiple path designs is to enable one or more of the following to contribute to the final classification: (a) different levels of abstraction, (b) different effective receptive fields, and (c) valuable information learned early to flow more easily to the classification stage.

In~\cite{He2015} (and subsequent extensions~\cite{Srivastava2015}\cite{Xie2017}\cite{Jgou2017}\cite{Zhang2020}) the authors added extra paths through the network with \textit{residual blocks} which are meta-layers that contained one or more convolutional operations as well as a ``skip connection'' that allowed information learned earlier in the network to skip over the convolutional operations.  Similarly, in~\cite{Szegedy2015a} and~\cite{Szegedy2015b}, the authors presented a network architecture that made heavy use of \textit{inception blocks}, which are meta-layers that branch from a previous layer into anywhere from 3 to 6 branches of varying layers of convolutions.  Then the branches were merged back together by concatenating the filters of those branches.  Let \(n\) be the average number of branches of different length (in terms of successive convolutions) and \(m\) be the number of successive inception blocks.  Then \(n\times m\) effective receptive fields and levels of abstraction are present at the output of the final inception block.  Additionally, the designs presented in both of these papers included two output stems (one branching out before going through additional inception blocks and the other after all inception blocks) each producing classification predictions.  These classifications were combined via static weighting to produce the final prediction.  In contrast to the aforementioned work, in this work, we present a network design that produces 3 output stems, each coming after a different number of convolutions, and \textit{thus representing different effective receptive fields and levels of abstraction}.  We conduct experiments that include statically weighted combinations as in~\cite{Szegedy2015a} and~\cite{Szegedy2015b}.  We then go further and investigate learning the branch weights \textit{simultaneously with all of the other network parameters via backpropagation}.  Again, in contrast to the aforementioned work, in these experiments, each of the separate classifications were performed with capsules rather than simple fully connected layers.

Our analysis of the existing literature shows that of the many branching methods explored, those that produced multiple final classifications merged those classifications via static weighting, which presupposes the relative importance of each output.  In this work we include and compare the results of both statically weighting the classification branches and learning the weights of the classification branches via backpropagation.

\subsection{Our Contribution}
Our contribution is as follows:
\begin{enumerate}
  \item We present a novel method for branching a CNN that allows for multiple effective receptive fields and levels of abstraction where each branch makes it's own classification prediction.  These classifications are then merged together, each contributing a ``vote''.  We present the results of experiments that include and compare both statically weighting the votes and learning the weights of the votes via backpropagation simultaneously with the rest of the network parameters.
  \item We do classification without any fully connected layers, but rather with HVCs.  HVCs are simpler, less computationally expensive, and our network design requires 5.5\(\times{}\) fewer parameters and 4\(\times{}\) fewer training epochs compared to the previously best performing capsule network, all while using no reconstruction sub-network and no computationally expensive routing mechanism.\@
  \item  This design, in combination with a domain-specific set of randomly applied augmentation techniques, establishes a new state of the art for the MNIST dataset with an accuracy of \textbf{99.87\%} for an ensemble of these models, as well as establishing a new state of the art for a single model (\textbf{99.83\%} accurate).
\end{enumerate}

\section{Proposed Network Design}\label{sec:proposed_network_design}

The starting point for the network design was a conventional convolutional neural network following many widely used practices.  These include stacked \(3\times3\) convolutions, each of which with ReLU~\cite{Glorot2011} activation preceded by batch normalization~\cite{Ioffe2015}.  We also followed the common practice of increasing the number of filters in each subsequent convolutional operation relative to the previous one.  Specifically, our first convolution uses 32 filters and each subsequent convolution uses 16 more filters than the previous one.  Additionally, the final operation before classification was to softmax the logits and to use categorical cross entropy for calculating loss.

One common design element found in many convolutional neural networks which we intentionally avoided was the use of any pooling operations.  We agree with Geoffrey Hinton's assessment~\cite{Hinton2018b} of pooling (a method of down-sampling) as an operation to be avoided due to the information it ``throws away''.  With the MNIST data being only \(28\times28\), we have no need to down-sample.  In choosing not to down-sample, we face the potential dilemma of how to reduce the dimensionality as we descend deeper into the network.  This dilemma is solved by choosing not to zero-pad the convolution operations and thus each convolution operation by its nature reduces the dimensionality by 2 in both the horizontal and vertical dimensions.  We deem choosing not to zero-pad as preferable in its own right in that zero padding effectively adds information not present in the original sample.

Rather than having a single monolithic design such that each operation in our network feeds into the next operation and only the next operation, we chose to create multiple branches.  After the first two sets of three convolutions, in addition to feeding to the subsequent convolution, we also branched off the output to be forwarded on to an additional operation (detailed next).  Thus, after all convolutions have been performed, we have three branches in our network.
\begin{enumerate}[label=\arabic*\rightparenthesis]
\item The first of which has been through three \(3\times{}3\) convolutions and consists of 64 filters each having an effective receptive field of 7 of the original image pixels.
\item The second of which has been through six \(3\times{}3\) convolutions and consists of 112 filters each having an effective receptive field of 11 of the original image pixels.
\item The third of which has been through nine \(3\times{}3\) convolutions and consists of 160 filters each having an effective receptive field of 15 of the original image pixels.
\end{enumerate}

\begin{figure}[!ht]
  \centering
  \includegraphics[width=0.35\textwidth]{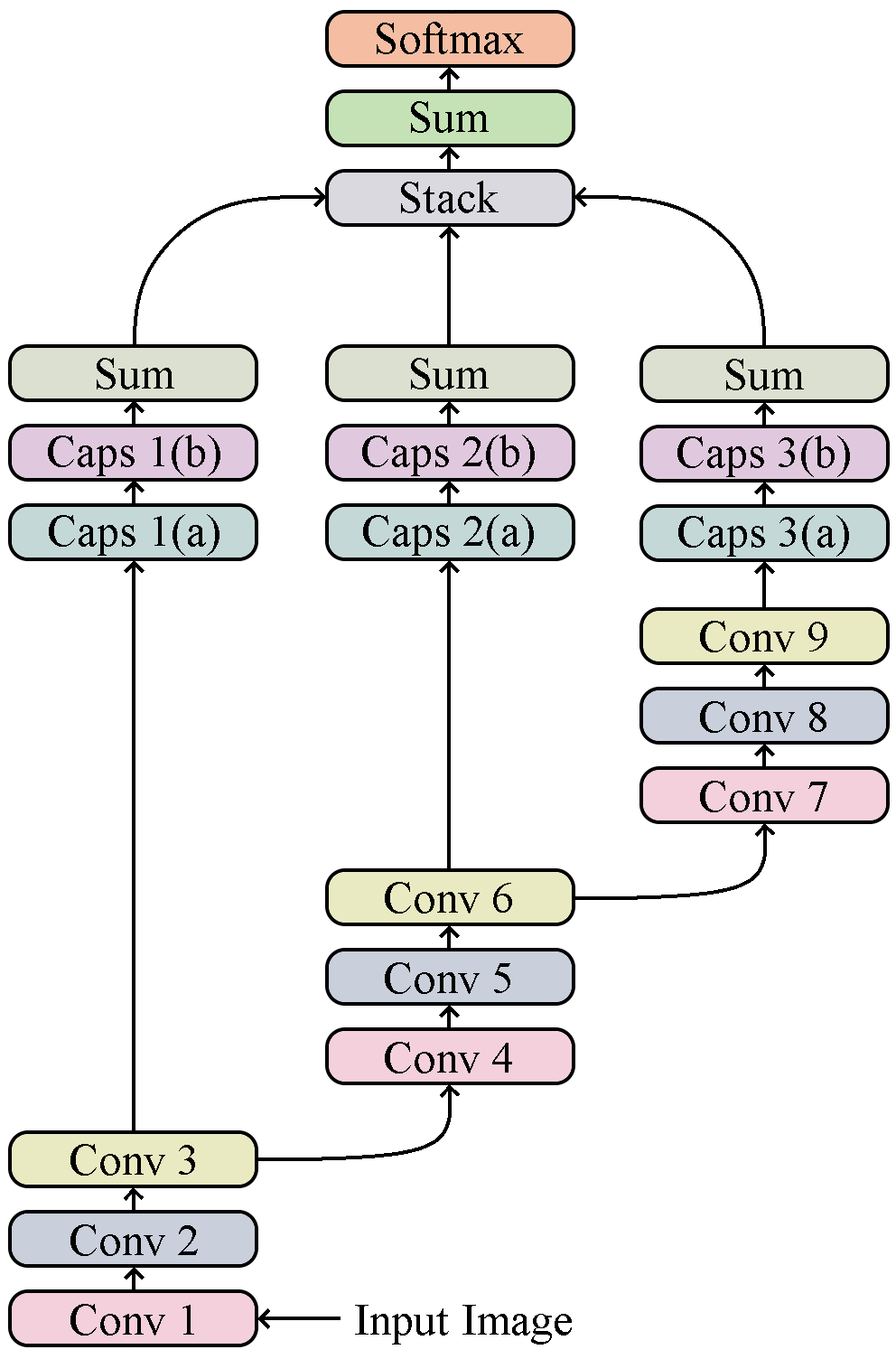}
  \caption{The proposed network from input to classification.}\label{fig:network_design_high_level}
\end{figure}

For each branch, rather than flattening the outputs of the convolutions into scalar neurons, we instead transformed each filter into a vector to form the first capsule in a pair of homogeneous vector capsules.  This operation is represented by ``Caps 1(a)'', ``Caps 2(a)'' and ``Caps 3(a)'' in \autoref{fig:network_design_high_level}.

We then performed element-wise multiplication of each of those capsules with a set of weight vectors (one for each class) of the same length.  This results in \textit{n}\(times{}\)\textit{m} weight vectors where \textit{n} is the number of capsules transformed from filter maps and \textit{m} is the number of classes.  We summed, per class (\textit{m}), each of the \textit{n} vectors to form the second capsule in each pair of homogeneous vector capsules.  After this that we applied batch normalization and then ReLU activation.  The process elucidated in this paragraph is represented by ``Caps 1(b)'', ``Caps 2(b)'' and ``Caps 3(b)'' in \autoref{fig:network_design_high_level}.

After the pairs of capsules for each breach, the second capsule vector in each pair is reduced to a single value per class by summing the components of the vector.  These values can be thought of as the branch-level logits.

Before classifying, the three branch-level sets of logits need to be reconciled with the fact that each image only belongs to one class.  This is accomplished by stacking each class's branch-level logits into vectors of length 3.  Then, each vector is reduced by summation to a single value to form the final set of logits to be classified from.  \autoref{fig:network_design_high_level} shows the high-level view of the entire network.

\begin{figure}[!ht]
  \centering
  \subfloat[In this example, the 4 filter maps have been converted into four 9-dimensional capsules, each made from an entire feature map.  The first 2 of 4 such capsules are highlighted in red and blue respectively.  For the sake of brevity, we will refer to this throughout the remainder of this work as using \textit{XY-Derived Capsules}.]{
  \begin{minipage}[c][1\width]{0.3\textwidth}
    \centering
    \includegraphics[width=0.6\textwidth]{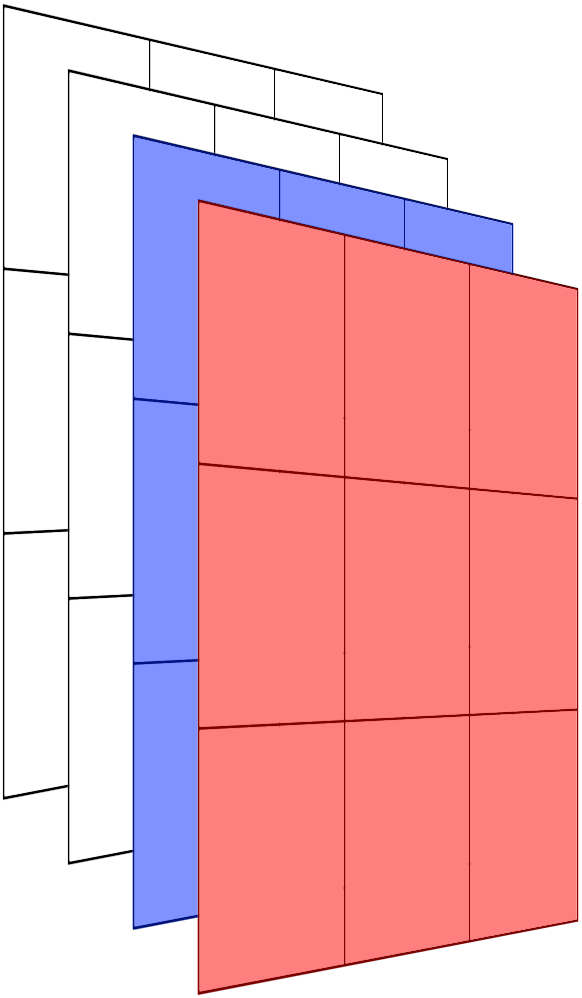}\label{fig:capsule_method_xy}
  \end{minipage}}
  \hspace*{.5in}
  \subfloat[In this example, the 4 filter maps have been converted into a single 4-dimensional capsule for each distinct \(x\) and \(y\) coordinate of the feature maps.  The first 2 of 9 such capsules are highlighted in red and blue respectively.  For the sake of brevity, we will refer to this throughout the remainder of this work as using \textit{Z-Derived Capsules}.]{
  \begin{minipage}[c][1\width]{0.3\textwidth}
    \centering
    \includegraphics[width=0.6\textwidth]{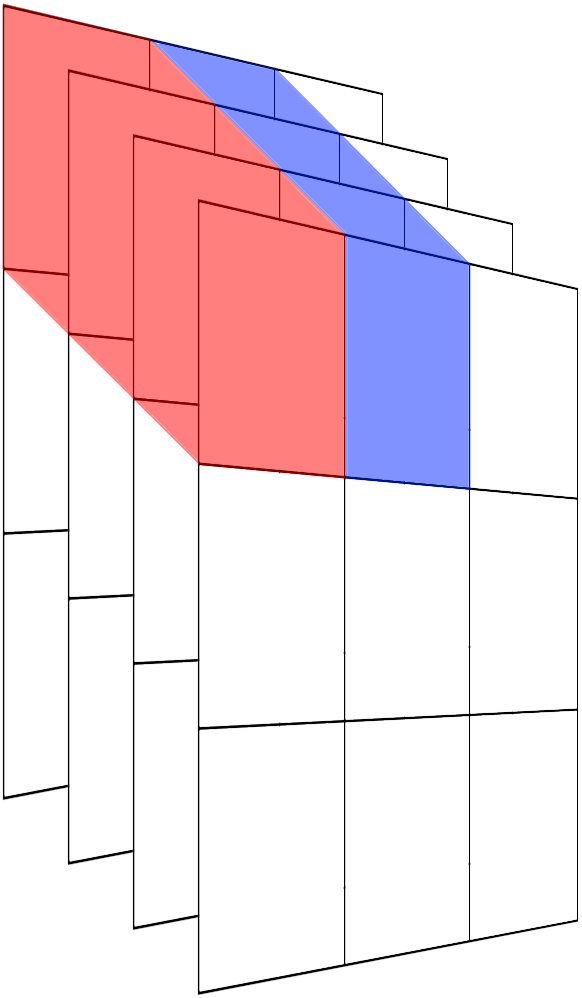}\label{fig:capsule_method_z}
  \end{minipage}}
  \caption{Illustrating the construction of capsules from 4 3\(\times{}\)3 filter maps.  These are the processes denoted by ``Caps 1(a)'', ``Caps 2(a)'', and ``Caps 3(a)'' in \autoref{fig:network_design_high_level}.}\label{fig:capsule_methods}
\end{figure}

In~\cite{Byerly2019}, we experimented with a variety of methods for constructing the first layer of capsules out of the preceding filter maps.  In this work, we limited our experiments to 2 of these methods (see \autoref{fig:capsule_methods}).  The first method constructs each capsule from each distinct feature map (a method that, for brevity, we will refer to as \textit{XY-Derived Capsules} in this work), whereas the second method constructs each capsule from each distinct \(x\) and \(y\) coordinate of the combination of all of the feature maps (a method that, for brevity, we will refer to as \textit{Z-Derived Capsules} in this work).

We used no weight decay regularization~\cite{Hinton1987}, a staple regularization method that improves generalization by penalizing the emergence of large weight values.  Nor did we use any form of dropout regularization~\cite{Hinton2012}\cite{Wan2013} which are regularization methods designed to stop the co-adaptation of weights.  We also did not use a reconstruction sub-network as in~\cite{Sabour2017}.  These decisions were made in order to investigate the generalization properties of our novel network design elements in the absence of other techniques associated with good generalization.  In addition, we intentionally left out any form of ``routing'' algorithm as in~\cite{Sabour2017} and~\cite{Hinton2018}, preferring to rely on traditional trainable weights and backpropagation.

\section{Experimental Setup}\label{sec:experimental_setup}

\subsection{Merge Strategies}\label{sec:experimental_setup_merge_strategies}

In~\cite{Szegedy2015a} and~\cite{Szegedy2015b}, the authors chose to give static, predetermined weights to both output branches and then added them together.  In our case, for both capsules configurations from \autoref{fig:capsule_methods}, we conducted three separate experiments of 32 trials each in order to investigate the effects of predetermined equal weighting of the branch outputs compared to learning the branch weights via backpropagation:

\begin{enumerate}[label=\arabic*\rightparenthesis]
\item \textbf{Not learnable}.  For this experiment, we merged the three branches together with equal weighting in order to investigate the effect of disallowing any one branch to have more impact than any other.
\item \textbf{Learnable with randomly initialized branch weights}. (Abbreviated as \textbf{Random Init.} subsequently.)  For this experiment, we allowed randomly initialized weights to be learned via backpropagation.
\item \textbf{Learnable with branch weights initialized to one}.  (Abbreviated as \textbf{Ones Init.} subsequently.)  For this experiment, we also allowed the weights to be learned via backpropagation.  The difference with the \textbf{Random Init.} experiment being that we initialized the weights to 1.  We conducted this experiment in addition to the \textbf{Random Init.} experiment in order to understand the difference between starting with random weights and starting with equal weights that are subsequently allowed to diverge during training.
\end{enumerate}

\subsection{Data Augmentation}\label{sec:experimental_setup_data_augmentation}

Most (but not all~\cite{Hasanpour2016}\cite{Chang2015}) of the state of the art MNIST results achieved over the past decade have used data augmentation~\cite{Sato2015}\cite{Wan2013}\cite{Ciresan2012}.  In addition to the network design, a major part of our work involved applying an effective data augmentation strategy that included transformations informed specifically by the domain of the data.  For example, we wanted to be sure we did not rotate our images into being more like a different class (e.g.\ rotating an image of the digit 2 by 180 degrees to create something that would more closely resemble a malformed 5).  Nor did we want to translate the image content off of the canvas and perhaps cut off the left side of an 8 and thus create a 3.  Choosing data augmentation techniques specific to the domain of interest is not without precedent (see for example~\cite{Ciresan2012} and~\cite{Sabour2017}, both of which used data augmentation techniques specific to MNIST).

By modern standards, in terms of dataset size, MNIST has a relatively low number of training images.  As such, judicious use of appropriate data augmentation techniques is important for achieving a high level of generalizability in a given model.  In terms of structure, hand-written digits show a wide variety in their rotation relative to some shared true ``north'', position within the canvas, width relative to their height, and the connectedness of the strokes used to create them.  Throughout training for all trials, every training image in every epoch was subjected to a series of four operations in order to simulate a greater variety of the values for these properties.

\begin{enumerate}[label=\arabic*\rightparenthesis] \item \textbf{Rotation}.  First, we randomly rotated each training image by up to 30 degrees in either direction.  Whether to actually apply this rotation was chosen by drawing from a Bernoulli distribution with probability p of 0.5 (a fair coin toss).
\item \textbf{Translation}.  Second, we randomly translated each training image within the available margin present in that image.  In~\cite{Sabour2017}, the authors limited their augmentation to shifting the training images randomly by up to 2 pixels in either or both directions.  The limit of only 2 pixels for the translation ensured that the translation is label-preserving.  As the MNIST training data has varying margins of non-digit space in the available \(28\times28\) pixel canvas, using more than 2 pixels randomly, would be to risk cutting off part of the digit and effectively changing the class of the image.  For example, a 7 that was shifted too far left could become more appropriately classed as a 1, or an 8 or 9 shifted far enough down could be more appropriately classed as a zero.  The highly structured nature of the MNIST training data allows for an algorithmic analysis of each image that will provide the translation range available for that specific image that will be guaranteed to be label-preserving.  \autoref{fig:example_mnist_digit} shows an example of an MNIST training image that has an asymmetric translation range that, as long as any translations are performed such that the digit part of the image is not moved by more pixels than are present in the margin, will be label preserving.  In other words, the specific training example shown in \autoref{fig:example_mnist_digit} could be shifted by up to 8 pixels to the left or 4 to the right and up to 5 up or 3 down, and after doing so, all of the pixels belonging to the actual digit will still be in the resulting translated image.  The amount within this margin to actually translate a training image was chosen randomly.  Whether to translate up or down and whether to translate left or right were drawn independently from a Bernoulli distribution with probability p of 0.5 (a fair coin toss).

\begin{figure}[ht]
  \centering
  \includegraphics[width=2.5in]{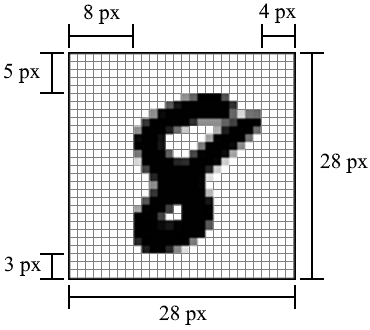}
  \caption{Example MNIST digit w/annotated margins.}\label{fig:example_mnist_digit}
\end{figure}

\item \textbf{Width}.  Third, we randomly adjusted each training image's width.  MNIST images are normalized to be within a \(20\times20\) central patch of the \(28\times28\) canvas.  This normalization is ratio-preserving, so all images are 20 pixels in the height dimension but vary in the number of pixels in the width dimension.  This variance not only occurs across digits, but intra-class as well, as different peoples' handwriting can be thinner or wider than average.  In order to train on a wider variety of these widths, we randomly compressed each image's width and then added equal zero padding on either side, leaving the digit's center where it was prior.  This was inspired by a similar approach adopted in~\cite{Ciresan2012}.  In our work, we compressed the width of each sample randomly within a range of 0--25\%.
\item \textbf{Random Erasure}.  Fourth, we randomly erased (setting to 0) a \(4\times4\) grid of pixels chosen from the central \(20\times20\) grid of pixels in each training image.  The X and Y coordinates of the patch were drawn independently from a random uniform distribution.  This was inspired by the random erasing data augmentation method in~\cite{Zhong2017}.  The intention behind this method was to expose the model to a greater variety of (simulated) connectedness within the strokes that make up the digits.  An alternative interpretation would be to see this as a kind of feature-space dropout.
\end{enumerate}

\subsection{Training}

We followed the training methodology from~\cite{Byerly2019} and trained with the Adam optimizer~\cite{Kingma2014} using all of the default/recommended parameter values, including the base learning rate of 0.001.  Also, as in both~\cite{Byerly2019} and~\cite{Sabour2017}, we exponentially decayed the base learning rate.  For our experiments, which trained for 300 epochs, we applied an exponential decay to the learning rate at a rate of 0.98 per epoch.

Test accuracy was measured using the exponential moving average of prior weights with a decay rate of 0.999.~\cite{Izmailov2018}

\section{Experimental Results}\label{sec:experimental_results}

\subsection{Individual Models}

For both of the capsule construction methods (see \autoref{fig:capsule_methods}) and each of the three merge strategies (see \autoref{sec:experimental_setup_merge_strategies}) we ran 32 trials.  Each trial had weights randomly initialized prior to training and, due to the stochastic nature of the data augmentation, a different set of training images.  As a result, training progressed to different points in the loss surface resulting in a range of values for the top accuracies that were achieved on the test set.  See \autoref{tab:individual_models}.

\begin{table}[!ht]
  \centering
  \begin{minipage}{\textwidth}
    \caption{Test Accuracy of the Individual Models}
    \begin{tabularx}{\textwidth}{@{}Xlrrrr@{}}
      \toprule
	      HVC Configuration & Experiment & Min & Max & Mean & SD \\
      \midrule
        \multirow{3}{*}{\begin{minipage}{2.5in}Using XY-Derived Capsules\end{minipage}}
	      & Not Learnable & 99.71\% & \textbf{99.79\%} & 0.997500 & 0.0002190 \\
	      & Random Init. & 99.72\% & \textbf{99.78\%} & 0.997512 & 0.0001499 \\
	      & Ones Init. & 99.70\% & 99.77\% & 0.997397 & 0.0001885 \\
      \midrule
        \multirow{3}{*}{\begin{minipage}{2.15in}Using Z-Derived Capsules\end{minipage}}
	      & Not Learnable & 99.74\% & \textbf{99.81\%} & 0.997731 & 0.0001825 \\
	      & Random Init. & 99.73\% & \textbf{99.80\%} & 0.997684 & 0.0002023 \\
	      & Ones Init. & 99.72\% & \textbf{99.83\%} & 0.997747 & 0.0002509 \\
      \bottomrule
    \end{tabularx}\\[0.05in]\label{tab:individual_models}
    \captionsetup{justification=justified,singlelinecheck=false}
    \caption*{In all cases, using the Z-Derived Capsules was superior to using the XY-Derived Capsules.  For Z-Derived Capsules, no merge strategy produced statistically significantly superior test accuracy.  For XY-Derived Capsules, the only statistically significant test accuracy result was that the Ones Init.\ strategy produced inferior accuracy.  It should be noted that, though no strategy produced statistically significantly superior \textit{test accuracies}, when branches were allowed to learn their weights, the weights learned were statistically significant.  (Bold indicates a surpassing of the previous state of the art for individual models on MNIST.)  (SD abbreviates Standard Deviation)}
  \end{minipage}
\end{table}

\subsection{Ensembles}

Ensembling multiple models together and predicting based on the majority vote among the ensembled models routinely outperforms the individual models' performances.  Ensembling can refer to either completely different model architectures with different weights or the same model architecture after being trained multiple times and finding different sets of weights that correspond to different locations in the loss surface.  The previous state of the art of 99.82\% was achieved using an ensemble of 30 different randomly generated model architectures~\cite{Kowsari2018}.  Our ensembling method used the same architecture but with different weights.  We calculated the majority vote of the predictions for all possible combinations of the weights produced by the 32 trials.  See \autoref{tab:ensembles}.

\begin{table}[!ht]
  \centering
  \begin{minipage}{\textwidth}
    \caption{Test Accuracy of the Ensembles}
    \begin{tabularx}{\textwidth}{@{}Xrrrrrr@{}}
      \toprule
	      & 99.87\% & 99.86\% & 99.85\% & 99.84\% & 99.83\% & 99.82\% \\
      \midrule
        \multicolumn{7}{l}{Using XY-Derived Capsules} \\
      \midrule
        \hspace{.2in} Not Learnable & 0 & 0 & 0 & 0 & 4 & 1,183 \\
	      \hspace{.2in} Random Init. & 0 & 0 & 0 & 0 & 21 & 2,069 \\
	      \hspace{.2in} Ones Init. & 0 & 0 & 0 & 1 & 19 & 1,292 \\
      \midrule
        \multicolumn{7}{l}{Using Z-Derived Capsules} \\
      \midrule
        \hspace{.2in} Not Learnable & 184 & 4,029 & 89,384 & 1,587,152 & 17,746,467 & 121,731,146 \\
        \hspace{.2in} Random Init. & 0 & 1,226 & 533,318 & 17,319,668 & 148,600,238 & 554,104,195 \\
        \hspace{.2in} Ones Init. & 64 & 9,920 & 1,113,217 & 34,635,994 & 426,947,909 & 1,279,126,811 \\
      \bottomrule
    \end{tabularx}\\[0.05in]\label{tab:ensembles}
    \captionsetup{justification=justified,singlelinecheck=false}
    \caption*{Shown here are the number of ensembles that were generated that either matched the previous state of the art of 99.82\% or exceeded it.}
  \end{minipage}
\end{table}

\subsection{Branch Weights}

What follows are visualizations of the final branch weights (after 300 epochs of training) for each of the branches in all 32 trials of the experiment wherein the branch weights were initialized to one for both HVC configurations.

In \autoref{fig:branch_weights_xy}, we see that for all trials, the ratio between the all three learned branch weights is consistent, demonstrating that the amount of contribution from each branch plays a significant role.  In \autoref{fig:branch_weights_z}, we see a similar, though less pronounced consistency between the first branch's weight and the other two branches, however, branches two and three show no significant difference.  Strikingly, when using XY-Derived Capsules we see that branch three (the one having gone through all nine convolutions) has learned to be a more significant contributor.  When using Z-Derived Capsules, branch one (the one having gone through only three convolutions) has learned to be a more significant contributor, but only slightly.  Indeed, in the latter configuration, the contributions from all three branches is much more equal.

The experiments with randomly initialized branch weights showed the same relative weight of the branches for the magnitude of the weights learned.  However, when the initial random branch weight was a negative number, it learned the negative value of that magnitude, and backpropagation took care of flipping the signs of weights as needed further up the network.

\begin{tabular}{@{}p{0.21in}p{2.70in}lp{0.05in}p{2.70in}l@{}}
  \multicolumn{3}{c}{\includegraphics[width=3in]{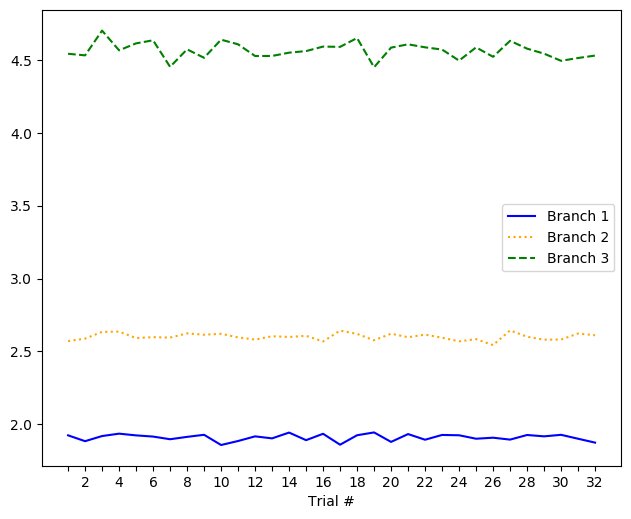}} &
  \multicolumn{3}{c}{\includegraphics[width=3in]{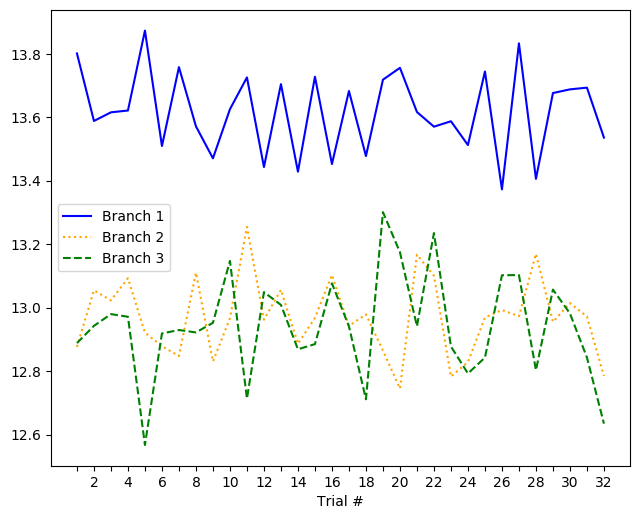}} \\[-0.1in]
  &
  \begin{figure_in_table}\caption{Final branch weights (after 300 epochs) for all 32 trials of the experiment using XY-Derived Capsules and for which the branch weights were initialized to one.}\label{fig:branch_weights_xy}\end{figure_in_table} &
  &
  &
  \begin{figure_in_table}\caption{Final branch weights (after 300 epochs) for all 32 trials of the experiment using Z-Derived Capsules and for which the branch weights were initialized to one.}\label{fig:branch_weights_z}\end{figure_in_table}
  &
  \\
\end{tabular}

Because the models using Z-Derived Capsules are clearly superior to XY-Derived Capsules, unless otherwise stated, all analyses throughout the remainder of this work will restrict attention to these 96 trials, and thus, when the text reads ``all 96 trials'', it should be understood that this refers to all 96 trials \textit{using Z-Derived Capsules}.

\subsection{Troublesome Digits}

Across all 96 trials there was total agreement on 9,912 out of the 10,000 test samples.  There were only 14 digits that were misclassified more often than not across all 96 trials.  This shows that although the accuracies of the models in the three experiments were quite similar, the different merge strategies of the three experiments did have a significant effect on classification.  Across all 96 trials, only 5 samples were misclassified in all models.  Those samples, as numbered by the order they appear in the MNIST test dataset (starting from 0) are 1901, 2130, 2597, 3422, and 6576.

\begin{figure}[!ht]
  \centering
  \setlength\tabcolsep{1pt}
  \begin{tabular}{@{}cccccc@{}}
    \includegraphics[width=0.4in]{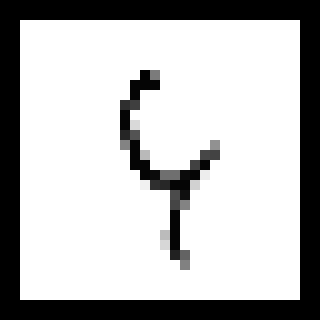} &
    \includegraphics[width=0.4in]{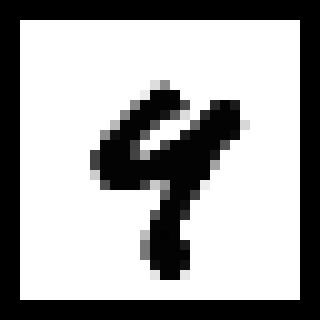} &
    \includegraphics[width=0.4in]{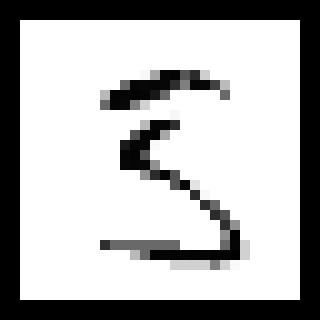} &
    \includegraphics[width=0.4in]{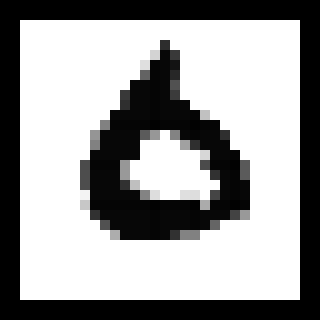} &
    \includegraphics[width=0.4in]{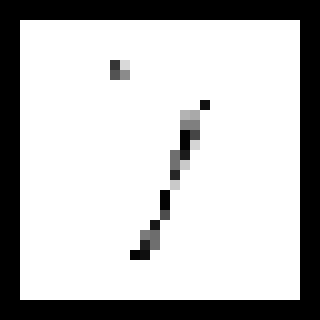} \\
    9 & 4 & 5 & 6 & 7 \\
    1901 & 2130 & 2597 & 3422 & 6576 \\
  \end{tabular}
  \caption{The Most Troublesome Digits}\label{fig:most_troublesome_digits}
\end{figure}

\subsection{MNIST State of the Art}

In \autoref{tab:previous_sota} we present a comparison of previous state of the art MNIST results for both single model evaluations and ensembles along with the results achieved in our experiments.
 
\begin{table}[!ht]
  \caption{Current and Previous MNIST State of the Art Results}
  \centering
  \begin{tabularx}{\textwidth}{@{}Xrr@{}}
    \toprule
      Paper & Year & Accuracy \\
    \midrule
      \multicolumn{3}{l}{Single Models} \\
    \midrule
      \hspace{.2in} Dynamic Routing Between Capsules\cite{Sabour2017} & 2017 & 99.75\% \\
      \hspace{.2in} Lets keep it simple, Using simple architectures to outperform deeper & 2016 & 99.75\% \\
      \hspace{.4in} and more complex architectures\cite{Hasanpour2016} \\
      \hspace{.2in} Batch-Normalized Maxout Network in Network\cite{Chang2015} & 2015 & 99.76\% \\
      \hspace{.2in} APAC:\@Augmented PAttern Classification with Neural Networks\cite{Sato2015} & 2015 & 99.77\% \\
      \hspace{.2in} Multi-Column Deep Neural Networks for Image Classification\cite{Ciresan2012} & 2012 & 99.77\% \\
      \hspace{.2in} \textbf{Using the method proposed in this work} & \textbf{2021} & \textbf{99.83\%} \\
    \midrule
      \multicolumn{3}{l}{Ensembles} \\
    \midrule
      \hspace{.2in} Regularization of Neural Networks using DropConnect\cite{Wan2013} & 2013 & 99.79\% \\
      \hspace{.2in} RMDL:\@Random Multimodel Deep Learning for Classification\cite{Kowsari2018} & 2018 & 99.82\% \\
      \hspace{.2in} \textbf{Using an ensemble of the method proposed in this work} & \textbf{2021} & \textbf{99.87\%} \\
    \bottomrule
  \end{tabularx}\label{tab:previous_sota}
\end{table}

How long a model takes to train is an important factor to consider when evaluating a neural network.  Indeed, it is an enabling factor during initial experimentation as faster training leads to a greater exploration of the design space.  In \autoref{tab:epochs_of_training} we present a comparison of the number of epochs of training used in experiments for the results achieved in the networks shown in \autoref{tab:previous_sota}.  Across all 96 trials, the design achieved peak accuracy in an average of 168 epochs, with a minimum peak achieved in 38 epochs and a maximum peak achieved at epoch 296.  Since, all trials were allowed to run for up to 300 epochs, that is the number reported in \autoref{tab:epochs_of_training}.

\begin{table}[!ht]
  \caption{Epochs of Training}
  \centering
  \begin{tabularx}{\textwidth}{@{}Xr@{}}
    \toprule
    	Paper & Epochs \\
    \midrule
      Dynamic Routing Between Capsules\cite{Sabour2017} & 1,200 \\
      APAC:\@Augmented PAttern Classification with Neural Networks\cite{Sato2015} & 15,000 \\
      Multi-Column Deep Neural Networks for Image Classification\cite{Ciresan2012} & 800 \\
      Regularization of Neural Networks using DropConnect\cite{Wan2013} & 1,200 \\
      RMDL:\@Random Multimodel Deep Learning for Classification\cite{Kowsari2018} & 120 \\
      \textbf{The method proposed in this work} & 300 \\
    \bottomrule
  \end{tabularx}\\[0.05in]\label{tab:epochs_of_training}
  \captionsetup{justification=justified,singlelinecheck=false}
  \caption*{Neither~\cite{Hasanpour2016} nor~\cite{Chang2015} report on how many epochs their designs were trained for.}
\end{table}

\subsection{Interpreting Capsules' Dimensions}

By adding a reconstruction sub-network to the overall network, it can be trained not just to classify the input digits, but also to reconstruct them.  Then, by following the method in~\cite{Sabour2017}, we can examine the effects of perturbing individual dimensions of the second set of capsules in a pair of HVCs.  The experiments using Z-Derived Capsules had capsules with 64, 112, and 160 dimensions.  When perturbing only one of that many dimensions the changes to the resulting constructed images are very subtle.  So we ran another experiment with no branches, reconstruction, and using multiple 8-dimensional capsules for each distinct \(x\) and \(y\) coordinate of the feature maps.  By perturbing one of only eight dimensions the effects are more visible and allows us to interpret the meaning of values in the digits' capsules (see \autoref{tab:dimensional_pertrubations}).

\begin{table}[!ht]
  \caption{Dimensional Perturbations}
  \centering
  \setlength\tabcolsep{1pt}
  \begin{tabularx}{\textwidth}{@{}Xm{3.25in}@{}}
    \toprule
    Rightward tilt & \includegraphics[width=3.25in]{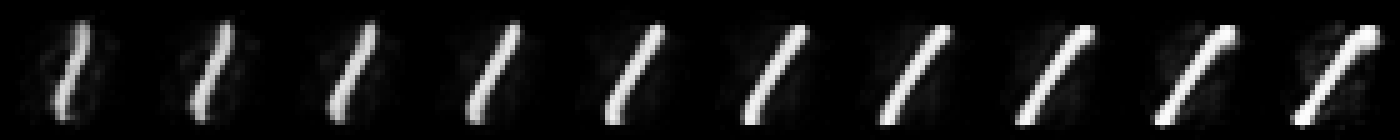} \\
    \midrule
      Top curl and height of lower loop & \includegraphics[width=3.25in]{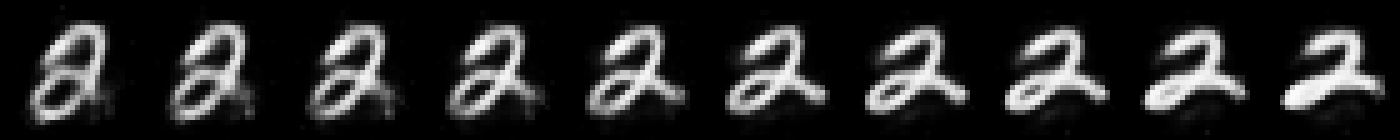} \\
    \midrule
      Length of lower stroke & \includegraphics[width=3.25in]{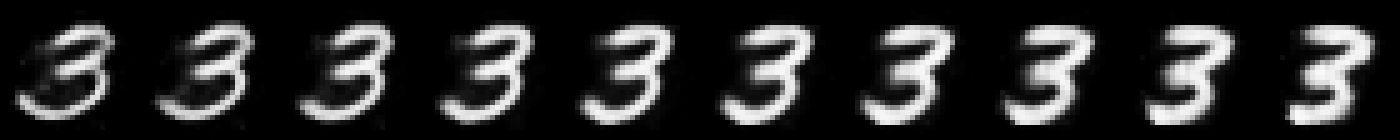} \\
    \midrule
      Angle of the top part of one stroke & \includegraphics[width=3.25in]{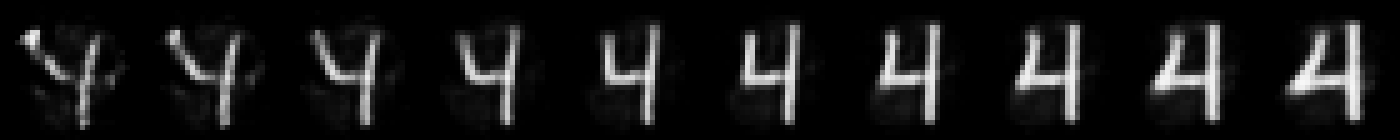} \\
    \midrule
      Sharpness of the angle of the lower two curves & \includegraphics[width=3.25in]{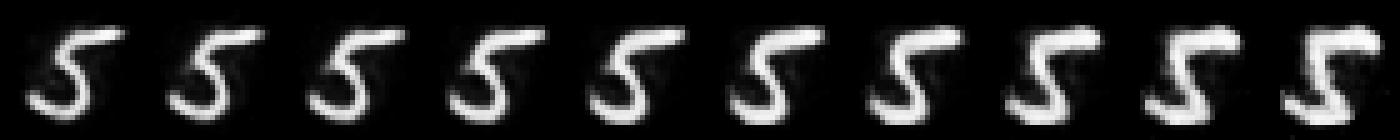} \\
    \midrule
      Width of entire digit & \includegraphics[width=3.25in]{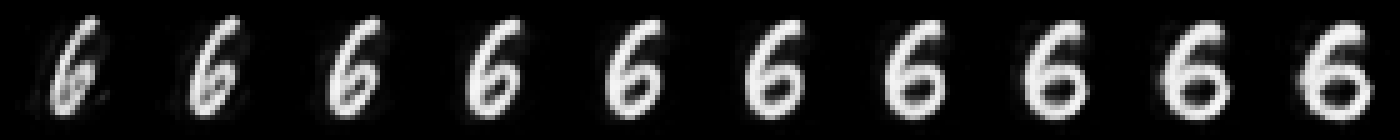} \\
    \midrule
      ``Hook'' in the initial part of the stroke & \includegraphics[width=3.25in]{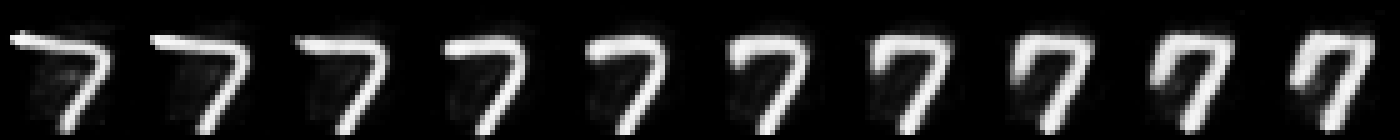} \\
    \midrule
      Width of lower loop & \includegraphics[width=3.25in]{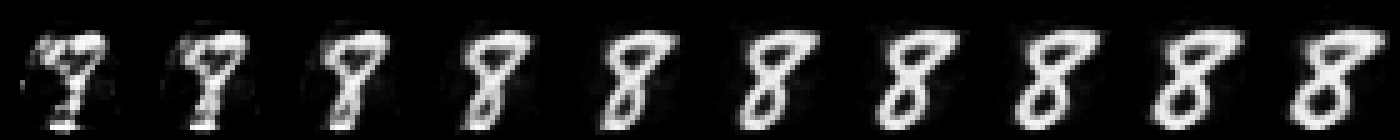} \\
    \midrule
      Lean angle & \includegraphics[width=3.25in]{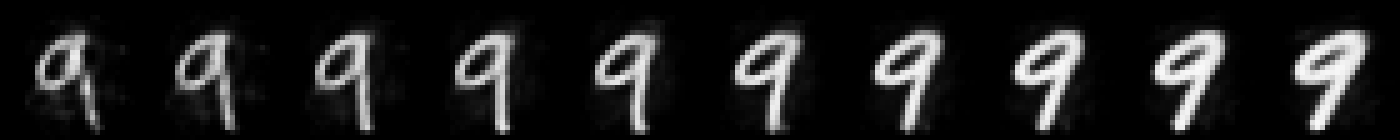} \\
    \bottomrule
  \end{tabularx}\\[0.05in]\label{tab:dimensional_pertrubations}
  \captionsetup{justification=justified,singlelinecheck=false}
  \caption*{Each row shows the reconstruction when one of the 8 dimensions in a digit's capsule is perturbed by intervals of 0.1 in the range [-0.5, 0.5].}
\end{table}

\subsection{Ablation Experiments}

In each of the following set of experiments, we compared the first 10 trials of the 32 trials for the Ones Init.\ merge strategy with 10 trials each of the additional experiments.

In~\cite{Sabour2017}, the authors used a custom loss function they called \textit{margin} loss combined with the mean squared error of the difference between the input images and the result of reconstructing them.  In our work and with our design, we chose to rely solely on categorical cross-entropy and not to use a reconstruction loss, as reconstruction adds a considerable number of parameters to the model (2.1M).  We ran two additional experiments to understand the effect of our choice of loss strategy (which used categorical cross-entropy and no reconstruction).  The first used margin loss and reconstruction, and the second used categorical cross-entropy and reconstruction.  There was no statistically significant difference among the three loss methods (see \autoref{tab:loss_comparison}).

\begin{table}[!ht]
  \caption{Comparison of Loss Methods}
  \centering
  \begin{tabularx}{\textwidth}{@{}Xrr@{}}
    \toprule
    	Loss Method & Mean Accuracy & SD \\
    \midrule
      Categorical Cross-Entropy (no reconstruction) & 99.7741\% & 0.000186455 \\
      Categorical Cross-Entropy (with reconstruction) & 99.7740\% & 0.000245764 \\
      Margin Loss (with reconstruction) & 99.7820\% & 0.000198997 \\
    \bottomrule
  \end{tabularx}\label{tab:loss_comparison}
\end{table}

In order to understand the relative importance of using HVCs vs.\ a fully connected layer and 3 branches vs.\ a single branch, we ran a series of experiments that ablated these components of the architecture.  \autoref{tab:network_structures} shows that HVCs are statistically significantly superior to a fully connected layer for both 1 and 3 branches, and shows that 3 branches are superior to 1 branch for both HVCs and a fully connected layer.

\begin{table}[!ht]
  \caption{Comparison of Network Structures}
  \centering
  \begin{tabularx}{\textwidth}{@{}Xrr@{}}
    \toprule
    	Network Structure & Mean Accuracy & SD \\
    \midrule
      Using HVCs and 3 branches & 99.7741\% & 0.000186455 \\
      Using HVCs and 1 branch & 99.7140\% & 0.000185472 \\
      Using a fully connected layer and 3 branches & 99.7550\% & 0.000111803 \\
      Using a fully connected layer and 1 branch & 99.6870\% & 0.000141774 \\
    \bottomrule
  \end{tabularx}\label{tab:network_structures}
\end{table}

In~\cite{Sabour2017}, the authors used translation, by a maximum of 2-pixels, as the only data augmentation method.  In our work, we devised a method for translating by up to the full margin available in any given direction.  We compared the effect of using only 2-pixel translation, only maximum margin translation, and our full suite of data augmentation methods.  Using the full suite of data augmentation methods was shown to be statistically superior to either of the other two methods.  Much to our surprise, we found that the 2-pixel translation method just barely crossed the threshold of being statistically significantly superior to the full margin translation method (see \autoref{tab:augmentation_strategies}).

The result we obtained by when using 2-pixel translation as the only data augmentation strategy allows for a direct comparison to the work of~\cite{Sabour2017}.  We obtained the same level of accuracy as they did, but using 5.5\(\times{}\) fewer parameters, 4\(\times{}\) fewer training epochs, no reconstruction sub-network, and requiring no routing mechanism.

\begin{table}[!ht]
  \caption{Comparison of Data Augmentation Strategies}
  \centering
  \begin{tabularx}{\textwidth}{@{}Xrr@{}}
    \toprule
    	Data Augmentation Strategy & Mean Accuracy & SD \\
    \midrule
      Translation (full margin), rotation, width adjustment, and random erasure & 99.7741\% & 0.000186455 \\
      Translation only (max. 2 pixels) (as in~\cite{Sabour2017}) & 99.7570\% & 0.000195192 \\
      Translation only (using full margin) & 99.7430\% & 0.000118743 \\
    \bottomrule
  \end{tabularx}\label{tab:augmentation_strategies}
\end{table}

\subsection{Additional Datasets}

In order to better understand the effect of the Z-Derived HVCs and additional branches, we ran additional sets of paired experiments for several additional datasets wherein the first set of experiments in a pair used the network design as described in this work and the second set of experiments excluded the Z-Derived HVCs and additional branches.  These second sets of experiments thus use a very small and typical convolutional neural network with 9 3\(\times{}\)3 convolutions and a final fully connected layer.

For MNIST and Fashion-MNIST we used the data augmentation strategy discussed in \autoref{sec:experimental_setup_data_augmentation}.  For CIFAR-10 and CIFAR-100, this data augmentation strategy is inappropriate, so we used a very typical strategy of randomly flipping the images horizontally and applying random adjustments to brightness, contrast, hue, and saturation.

For all four datasets, the model that included Z-Derived HVCs and 3 branches achieved the higher mean accuracy with statistical significance (see \autoref{tab:other_datasets}).

The fact that the accuracies for Fashion-MNIST~\cite{Xiao2017}, CIFAR-10, and CIFAR-100~\cite{Krizhevsky2009} were not competitive with current state of the art for those datasets is not especially surprising for several reasons.  First, our network was designed for optimal accuracy on classification of Arabic numerals which are highly structured and significantly simpler than the types of data in the other three datasets.  Second, due to the significantly simpler nature of MNIST, we used a small number of parameters for our network (1.5M).  For comparison, models competitive with state of the art for CIFAR-10 and CIFAR-100 use 10s and even 100s of millions of parameters.  Finally, models competitive with state of the art for CIFAR-10 and CIFAR-100 use additional training data beyond the canonical set for each, and we used no additional training data.

\begin{table}[!ht]
  \caption{Effects of Z-Derived HVCs and Branching on Additional Datasets}
  \centering
    \begin{tabularx}{\textwidth}{@{}lXrrrr@{}}
      \toprule
	      Dataset & Network Architecture & Max & Mean & SD & p-value\\
      \midrule
        \multirow{2}{*}{MNIST}
        & Z-Derived HVCs and 3 Branches & \textbf{99.81\%} & \textbf{99.7741\%} & 0.0001864 &
        \multirow{2}{*}{\num{1.824e-7}} \\
        & A Fully Connected Layer and 1 Branch & 99.71\% & 99.6870\% & 0.0001417 \\
      \midrule
        \multirow{2}{*}{Fashion-MNIST}
        & Z-Derived HVCs and 3 Branches & \textbf{93.89\%} & \textbf{93.6850\%} & 0.0016391 &
        \multirow{2}{*}{\num{5.243e-6}} \\
        & A Fully Connected Layer and 1 Branch & 93.36\% & 93.0410\% & 0.0014616 \\
      \midrule
        \multirow{2}{*}{CIFAR-10}
        & Z-Derived HVCs and 3 Branches & \textbf{89.23\%} & \textbf{88.9290\%} & 0.0015514 &
        \multirow{2}{*}{0.020898} \\
        & A Fully Connected Layer and 1 Branch & 89.06\% & 88.7500\% & 0.0017515 \\
      \midrule
        \multirow{2}{*}{CIFAR-100}
        & Z-Derived HVCs and 3 Branches & \textbf{64.15\%} & \textbf{63.8260\%} & 0.0026743 &
        \multirow{2}{*}{\num{6.859e-6}} \\
        & A Fully Connected Layer and 1 Branch & 62.96\% & 62.3760\% & 0.0035046 \\
      \bottomrule
    \end{tabularx}\\[0.05in]\label{tab:other_datasets}
    \captionsetup{justification=justified,singlelinecheck=false}
    \caption*{MNIST results come from the same experiments detailed in \autoref{tab:network_structures} and are repeated here to facilitate ease of comparison.  We conducted 10 trials of each unique type of experiment in order to establish statistical significance.}
\end{table}

\section{Conclusion}\label{sec:conclusion}

In this work, we proposed using a simple convolutional neural network and established design principles as a basis for a network architecture.  We then presented a design that branched out of the series of stacked convolutions at different points to capture different levels of abstraction and effective receptive fields, and from these branches, rather than flattening to individual scalar neurons, used Homogeneous Vector Capsules instead.

We also investigated three different methods of merging the output of the branches back into a single set of logits.  Each of the three merge strategies generated models that could be ensembled to create new state of the art results.

Beyond the network architecture, we proposed a robust and domain specific data augmentation strategy aimed at simulating a wider variety of renderings of the digits.

In doing this work, we established new MNIST state of the art accuracies for both a single model and an ensemble.  In addition to the network design and augmentation strategy, the ability to use an adaptive gradient descent method~\cite{Byerly2019} allowed us to achieve this on consumer hardware (2x NVIDIA GeForce GTX 1080 Tis in an otherwise unremarkable workstation) and was an enabling factor in both initial explorations and the training of all 322 trials of experiments referenced in this work.

\printbibliography{}

\bigskip

\begin{tabularx}{\textwidth}{@{}l@{}}
  The code used for all experiments and summary level data is publicly available on GitHub at: \\
  \href{https://github.com/AdamByerly/BMCNNwHFCs}{https://github.com/AdamByerly/BMCNNwHFCs} \\
\end{tabularx}

\bigskip

\appendix

\section{Appendix}\label{sec:appendix}

\subsection{Digits Disagreed Upon}

What follows is the complete set of 88 digits that were predicted correctly by at least one model and incorrectly by at least one model.  These in combination with the digits from \autoref{fig:most_troublesome_digits} represent the complete set of digits that were not predicted correctly by all 96 trials.  Each image is captioned first by the class label in the test data set associated with the image, then the number of trials that predicted it correctly, and last the index of the digit in the test data. For example, the first image presented below has a class label of 3, 95 trials predicted that correctly, and it exists at index 87 in the MNIST test data.

\vspace{0.15in}

\centering
\setlength\tabcolsep{1pt}
\begin{tabular}{@{}ccccccccccccccc@{}}
  \includegraphics[width=0.4in]{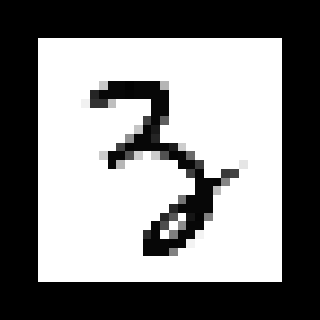} &
  \includegraphics[width=0.4in]{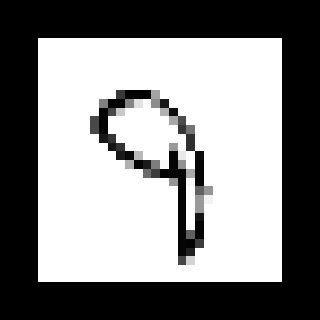} &
  \includegraphics[width=0.4in]{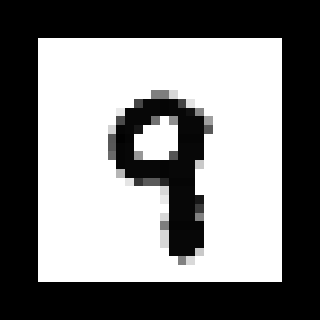} &
  \includegraphics[width=0.4in]{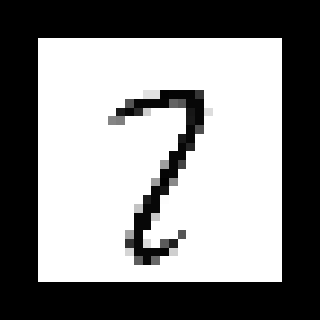} &
  \includegraphics[width=0.4in]{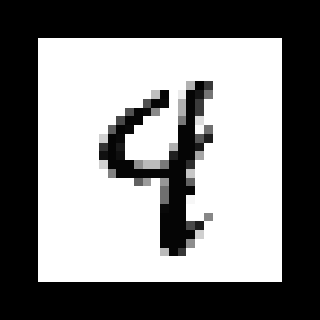} &
  \includegraphics[width=0.4in]{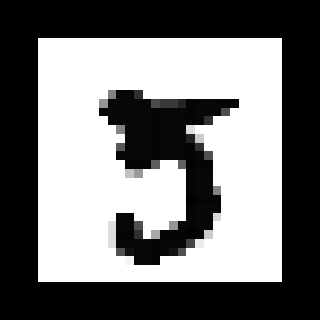} &
  \includegraphics[width=0.4in]{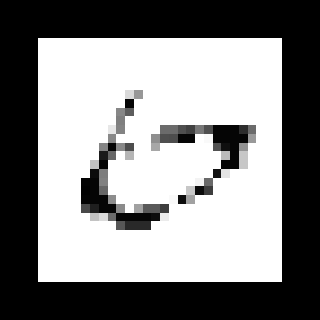} &
  \includegraphics[width=0.4in]{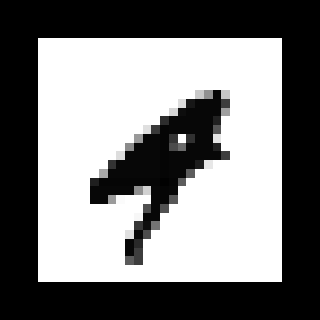} &
  \includegraphics[width=0.4in]{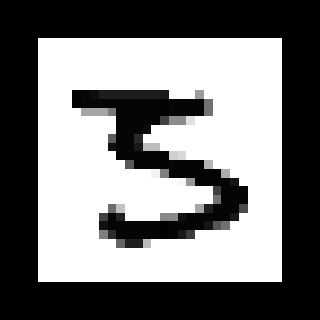} &
  \includegraphics[width=0.4in]{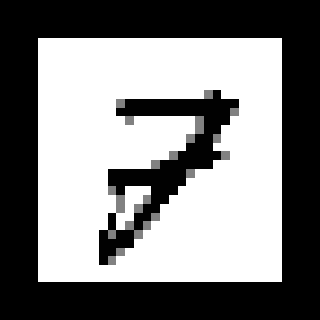} &
  \includegraphics[width=0.4in]{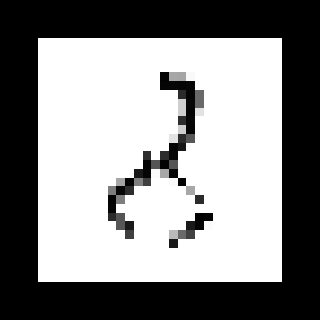} &
  \includegraphics[width=0.4in]{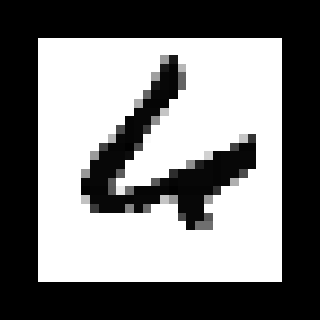} &
  \includegraphics[width=0.4in]{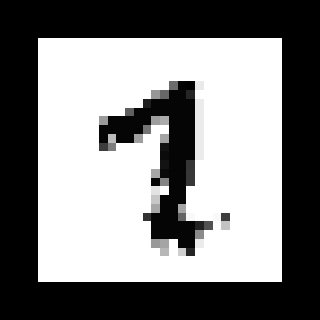} &
  \includegraphics[width=0.4in]{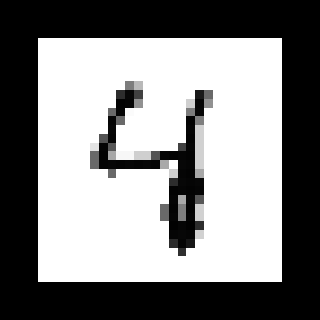} &
  \includegraphics[width=0.4in]{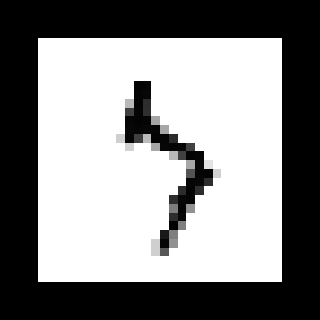} \\
  3 & 9 & 9 & 2 & 9 & 5 & 6 & 4 & 3 & 7 & 8 & 6 & 2 & 4 & 7 \\
  95 & 6 & 95 & 24 & 85 & 70 & 92 & 74 & 94 & 95 & 95 & 66 & 95 & 93 & 95 \\
  87 & 193 & 214 & 321 & 359 & 412 & 445 & 447 & 449 & 468 & 582 & 625 & 659 & 708 & 726 \\[.1 in]
\end{tabular}
\begin{tabular}{@{}ccccccccccccccc@{}}
  \includegraphics[width=0.4in]{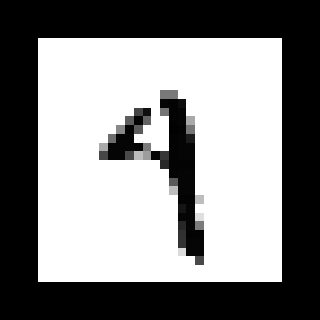} &
  \includegraphics[width=0.4in]{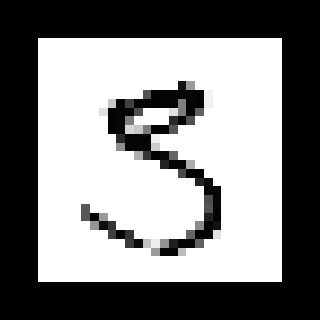} &
  \includegraphics[width=0.4in]{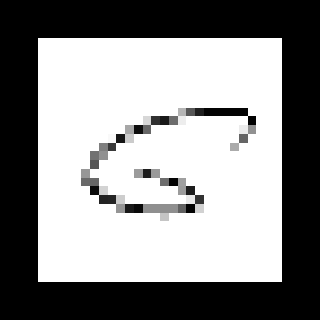} &
  \includegraphics[width=0.4in]{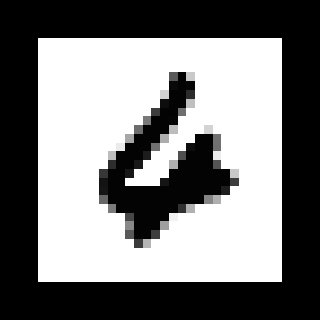} &
  \includegraphics[width=0.4in]{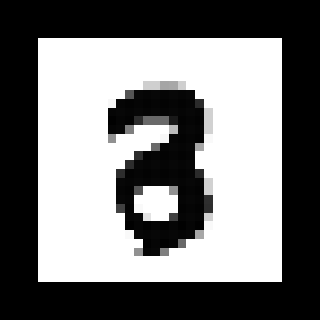} &
  \includegraphics[width=0.4in]{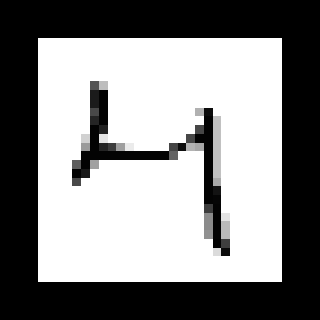} &
  \includegraphics[width=0.4in]{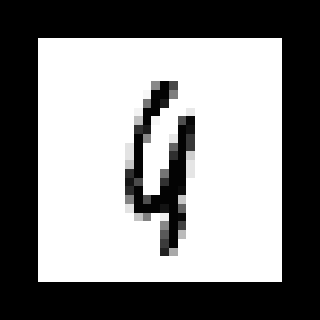} &
  \includegraphics[width=0.4in]{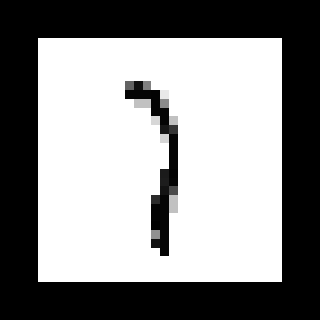} &
  \includegraphics[width=0.4in]{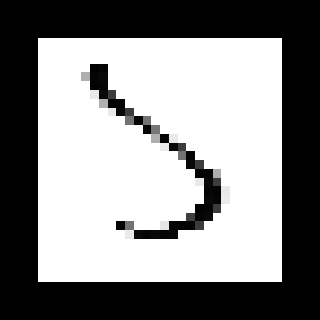} &
  \includegraphics[width=0.4in]{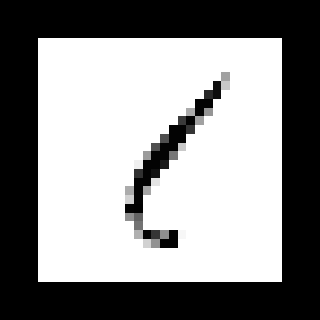} &
  \includegraphics[width=0.4in]{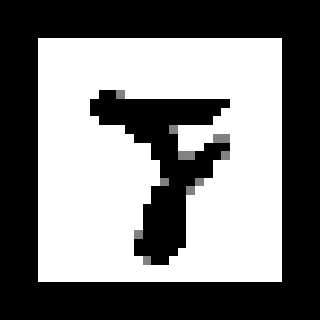} &
  \includegraphics[width=0.4in]{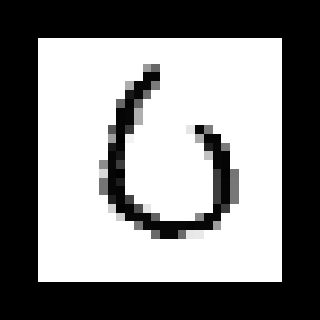} &
  \includegraphics[width=0.4in]{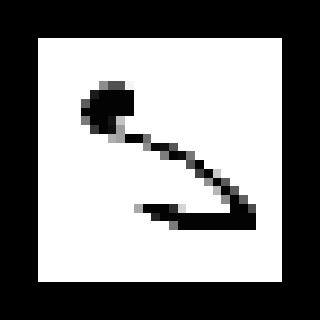} &
  \includegraphics[width=0.4in]{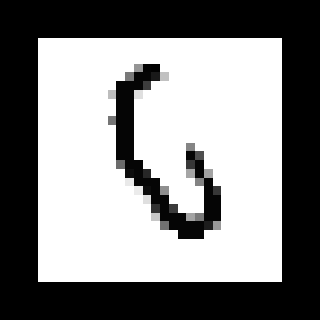} &
  \includegraphics[width=0.4in]{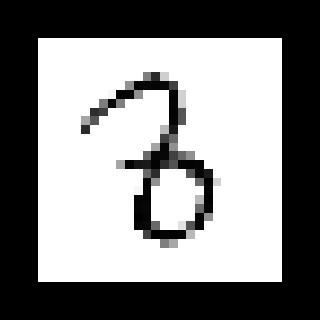} \\
  4 & 3 & 6 & 4 & 3 & 4 & 9 & 7 & 5 & 1 & 8 & 0 & 5 & 6 & 8 \\
  95 & 95 & 91 & 58 & 95 & 95 & 18 & 94 & 38 & 91 & 95 & 95 & 66 & 91 & 94 \\
  740 & 938 & 1014 & 1112 & 1114 & 1147 & 1232 & 1260 & 1393 & 1403 & 1530 & 1621 & 1737 & 1822 & 1878 \\[.1 in]
\end{tabular}
\begin{tabular}{@{}ccccccccccccccc@{}}
  \includegraphics[width=0.4in]{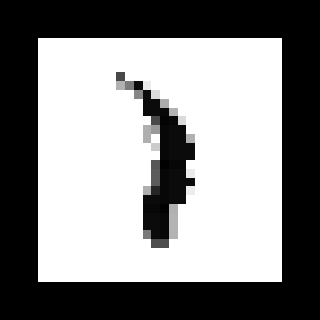} &
  \includegraphics[width=0.4in]{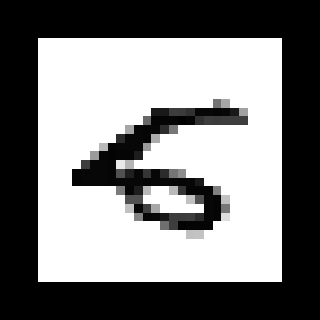} &
  \includegraphics[width=0.4in]{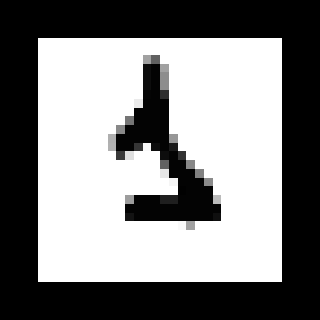} &
  \includegraphics[width=0.4in]{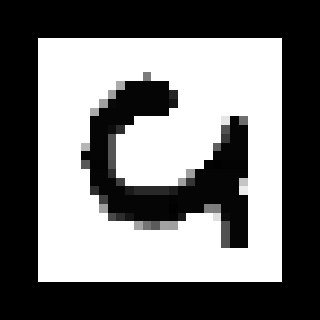} &
  \includegraphics[width=0.4in]{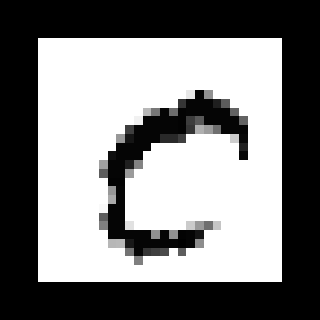} &
  \includegraphics[width=0.4in]{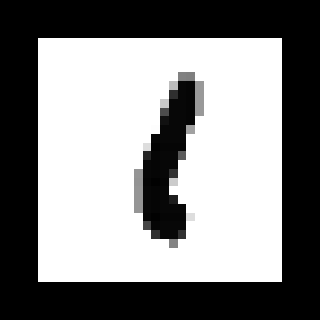} &
  \includegraphics[width=0.4in]{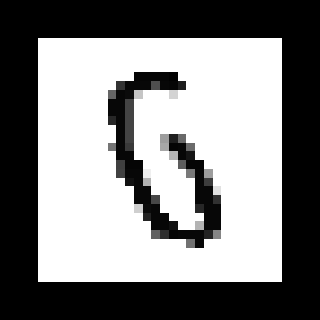} &
  \includegraphics[width=0.4in]{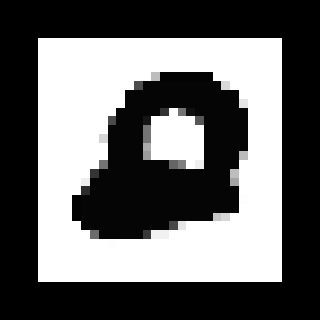} &
  \includegraphics[width=0.4in]{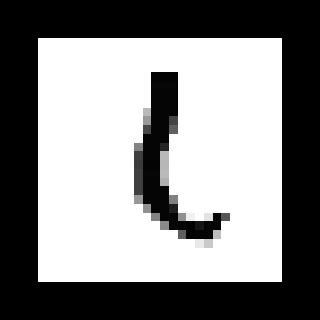} &
  \includegraphics[width=0.4in]{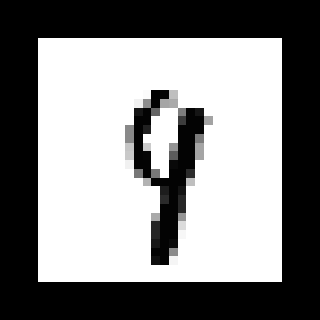} &
  \includegraphics[width=0.4in]{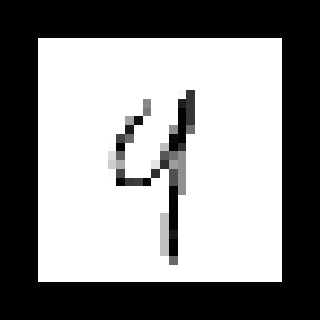} &
  \includegraphics[width=0.4in]{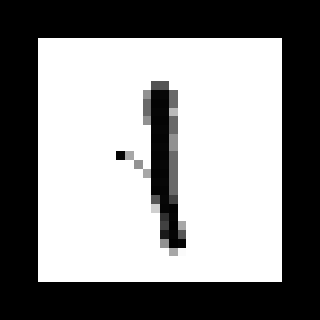} &
  \includegraphics[width=0.4in]{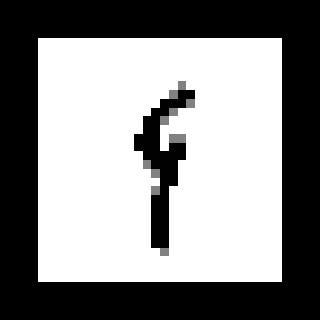} &
  \includegraphics[width=0.4in]{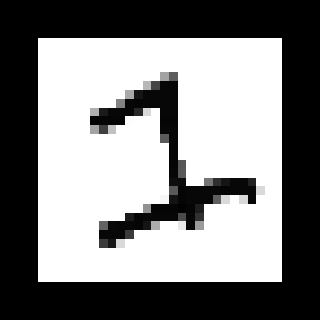} &
  \includegraphics[width=0.4in]{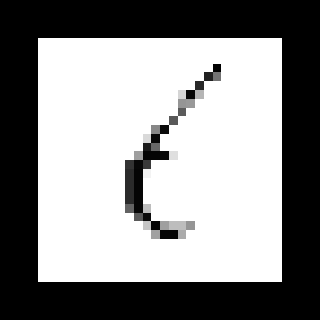} \\
  1 & 5 & 1 & 9 & 0 & 1 & 6 & 2 & 6 & 9 & 4 & 1 & 9 & 1 & 6 \\
  94 & 85 & 95 & 22 & 86 & 90 & 54 & 72 & 74 & 85 & 20 & 95 & 92 & 93 & 95 \\
  2018 & 2040 & 2266 & 2293 & 2326 & 2355 & 2454 & 2462 & 2654 & 2720 & 2771 & 2803 & 3005 & 3073 & 3365 \\[.1 in]
\end{tabular}
\begin{tabular}{@{}ccccccccccccccc@{}}
  \includegraphics[width=0.4in]{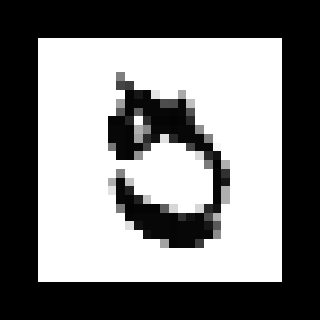} &
  \includegraphics[width=0.4in]{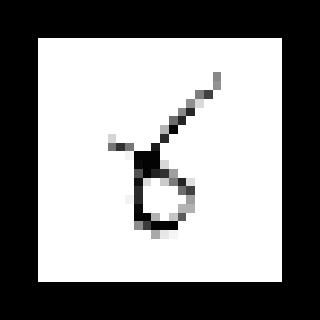} &
  \includegraphics[width=0.4in]{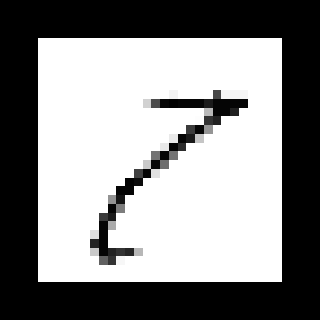} &
  \includegraphics[width=0.4in]{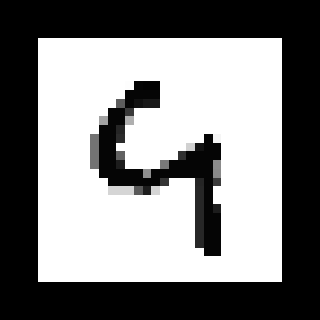} &
  \includegraphics[width=0.4in]{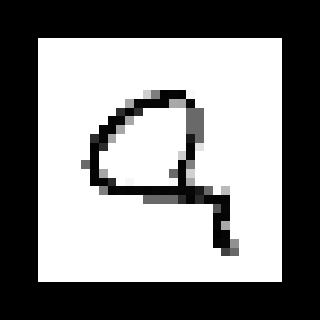} &
  \includegraphics[width=0.4in]{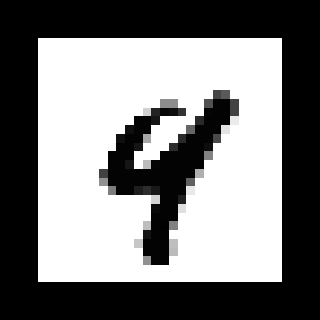} &
  \includegraphics[width=0.4in]{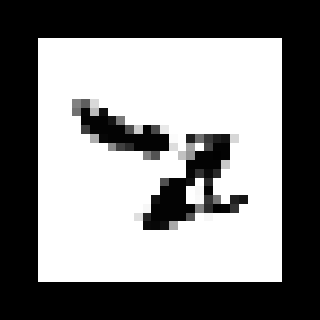} &
  \includegraphics[width=0.4in]{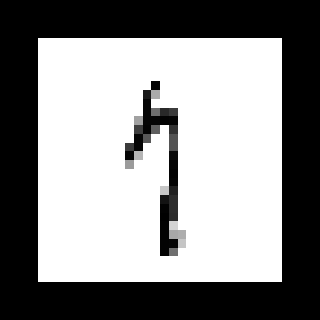} &
  \includegraphics[width=0.4in]{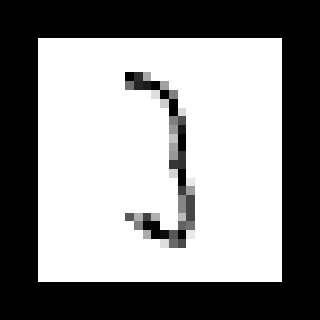} &
  \includegraphics[width=0.4in]{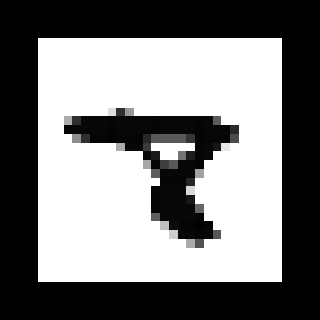} &
  \includegraphics[width=0.4in]{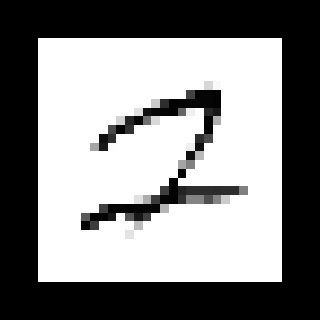} &
  \includegraphics[width=0.4in]{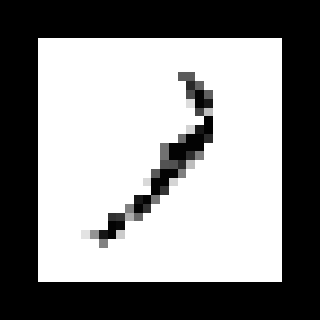} &
  \includegraphics[width=0.4in]{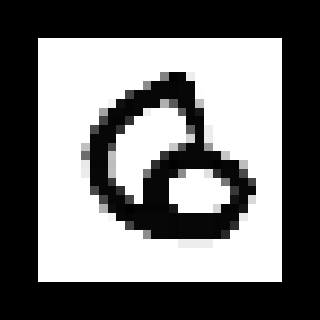} &
  \includegraphics[width=0.4in]{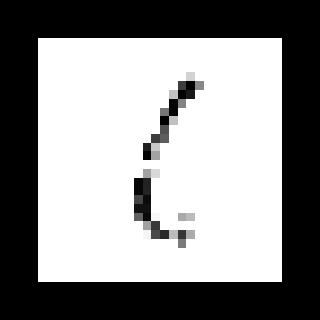} &
  \includegraphics[width=0.4in]{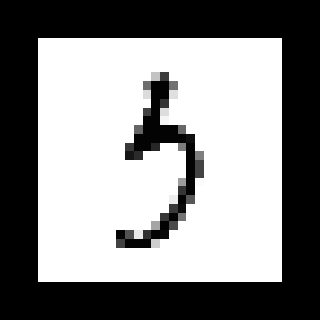} \\
  5 & 6 & 7 & 9 & 9 & 9 & 2 & 1 & 3 & 8 & 2 & 1 & 6 & 6 & 3 \\
  92 & 7 & 78 & 95 & 95 & 94 & 82 & 4 & 93 & 95 & 80 & 84 & 93 & 70 & 49 \\
  3558 & 3762 & 3808 & 3821 & 3859 & 3869 & 4176 & 4201 & 4443 & 4497 & 4504 & 4507 & 4571 & 4699 & 4740 \\[.1 in]
\end{tabular}
\begin{tabular}{@{}ccccccccccccccc@{}}
  \includegraphics[width=0.4in]{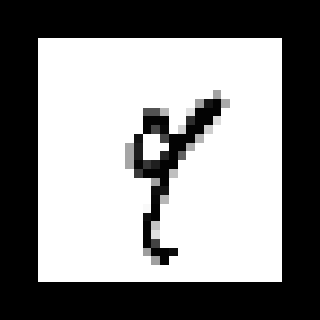} &
  \includegraphics[width=0.4in]{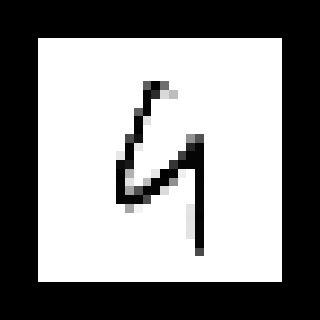} &
  \includegraphics[width=0.4in]{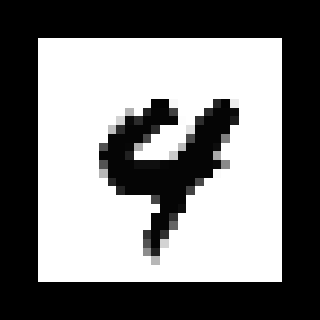} &
  \includegraphics[width=0.4in]{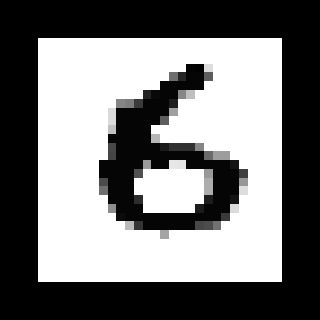} &
  \includegraphics[width=0.4in]{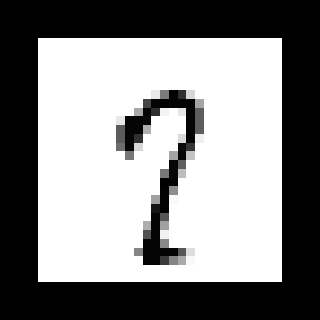} &
  \includegraphics[width=0.4in]{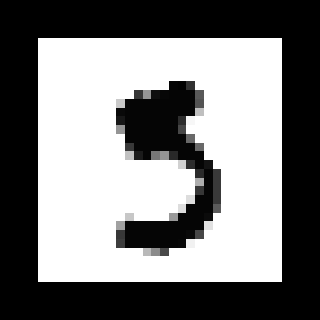} &
  \includegraphics[width=0.4in]{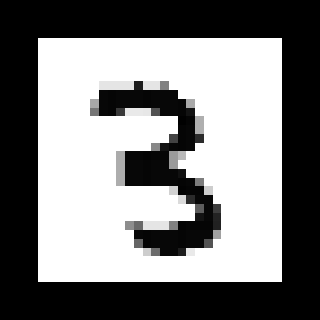} &
  \includegraphics[width=0.4in]{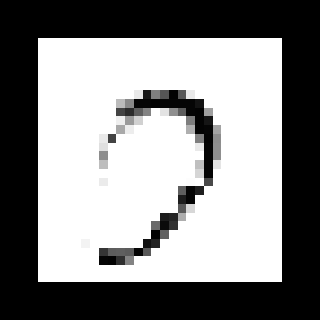} &
  \includegraphics[width=0.4in]{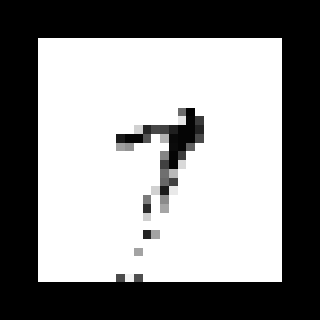} &
  \includegraphics[width=0.4in]{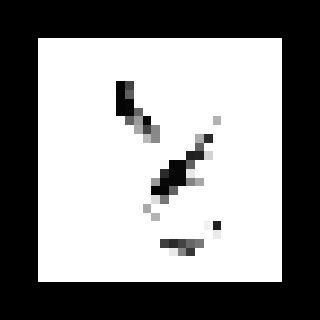} &
  \includegraphics[width=0.4in]{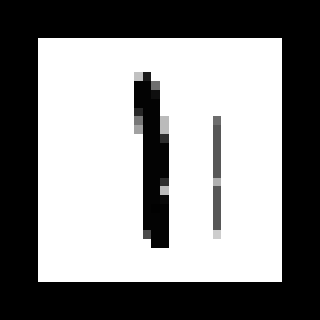} &
  \includegraphics[width=0.4in]{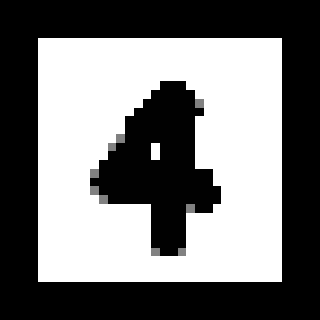} &
  \includegraphics[width=0.4in]{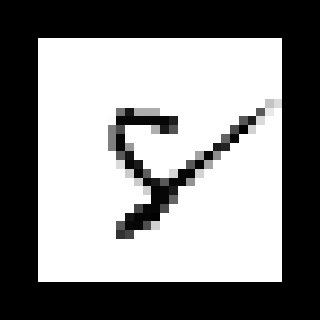} &
  \includegraphics[width=0.4in]{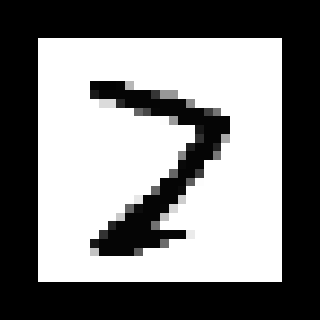} &
  \includegraphics[width=0.4in]{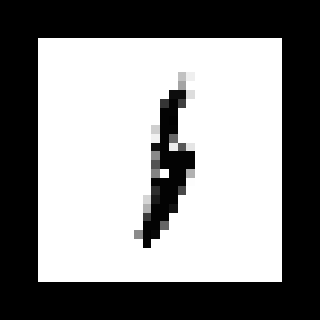} \\
  9 & 9 & 4 & 6 & 7 & 5 & 3 & 0 & 7 & 8 & 1 & 4 & 8 & 7 & 1 \\
  87 & 28 & 95 & 95 & 45 & 93 & 93 & 95 & 94 & 46 & 92 & 95 & 86 & 30 & 71 \\
  4761 & 4823 & 4860 & 4934 & 5654 & 5937 & 6371 & 6597 & 6599 & 6625 & 8020 & 8061 & 8279 & 8316 & 8376 \\[.1 in]
\end{tabular}
\begin{tabular}{@{}ccccccccccccc@{}}
  \includegraphics[width=0.4in]{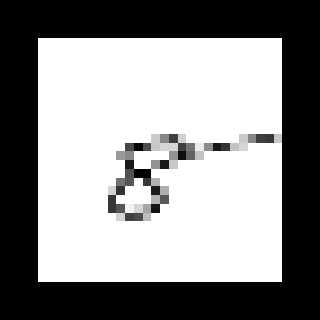} &
  \includegraphics[width=0.4in]{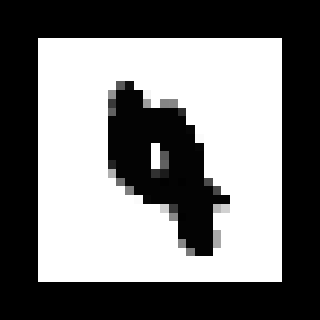} &
  \includegraphics[width=0.4in]{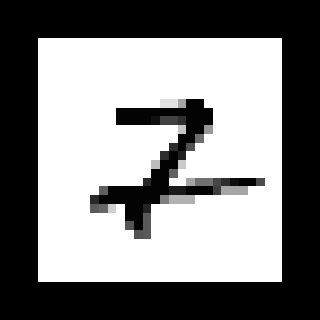} &
  \includegraphics[width=0.4in]{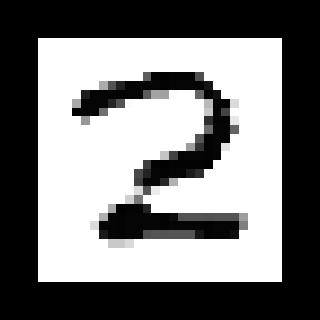} &
  \includegraphics[width=0.4in]{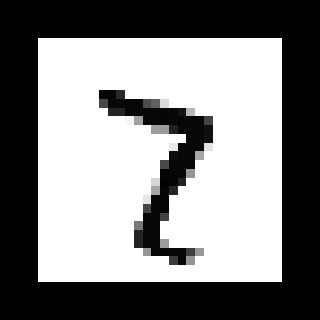} &
  \includegraphics[width=0.4in]{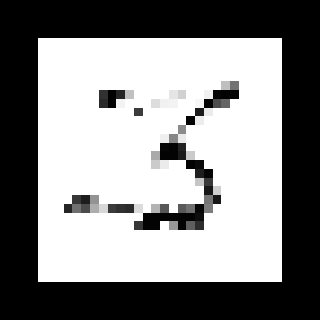} &
  \includegraphics[width=0.4in]{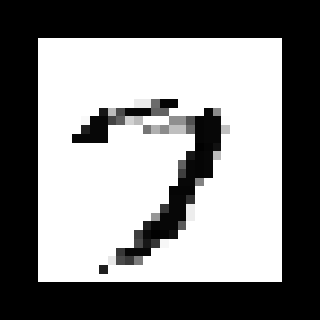} &
  \includegraphics[width=0.4in]{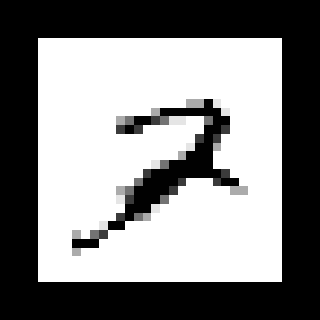} &
  \includegraphics[width=0.4in]{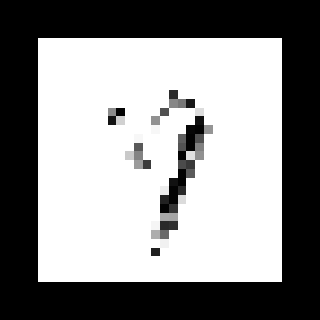} &
  \includegraphics[width=0.4in]{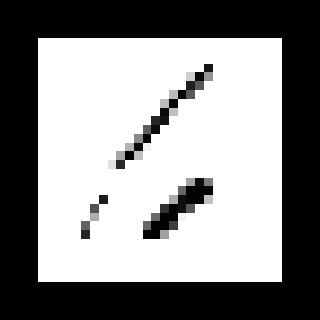} &
  \includegraphics[width=0.4in]{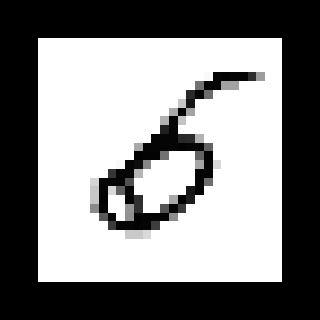} &
  \includegraphics[width=0.4in]{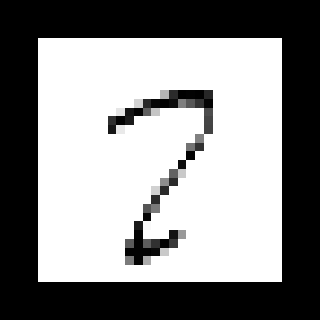} &
  \includegraphics[width=0.4in]{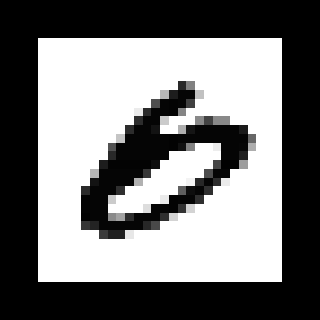} \\
  8 & 4 & 7 & 2 & 7 & 3 & 7 & 2 & 9 & 6 & 5 & 2 & 0 \\
  50 & 8 & 94 & 91 & 59 & 95 & 90 & 82 & 94 & 94 & 26 & 95 & 78 \\
  8408 & 8527 & 9015 & 9123 & 9505 & 9636 & 9637 & 9664 & 9692 & 9698 & 9729 & 9839 & 9850 \\[.1 in]
\end{tabular}

\end{document}